\documentclass[conference]{IEEEtran}
\IEEEoverridecommandlockouts
\usepackage{cite}
\usepackage{amsmath,amssymb,amsfonts}
\usepackage{algorithmic}
\usepackage{graphicx}
\usepackage{textcomp}
\usepackage[dvipsnames]{xcolor}
\usepackage{xspace}
\usepackage{multirow}
\usepackage{booktabs}
\usepackage{comment}

\def\BibTeX{{\rm B\kern-.05em{\sc i\kern-.025em b}\kern-.08em
    T\kern-.1667em\lower.7ex\hbox{E}\kern-.125emX}}
\begin{document}

\title{{SSR} : SAM is a Strong Regularizer for domain adaptive semantic segmentation\\
}

\newif\ifreview
\makeatletter
\DeclareRobustCommand\onedot{\futurelet\@let@token\@onedot}
\def\@onedot{\ifx\@let@token.\else.\null\fi\xspace}

\def\eg{\emph{e.g}\onedot} \def\Eg{\emph{E.g}\onedot}
\def\ie{\emph{i.e}\onedot} \def\Ie{\emph{I.e}\onedot}
\def\cf{\emph{c.f}\onedot} \def\Cf{\emph{C.f}\onedot}
\def\etc{\emph{etc}\onedot} \def\vs{\emph{vs}\onedot}
\def\wrt{w.r.t\onedot} \def\dof{d.o.f\onedot}
\def\etal{\emph{et al}\onedot}
\makeatother

%\reviewtrue

\ifreview
\author{\IEEEauthorblockN{Anonymous Authors}}
\else
\author{\IEEEauthorblockN{Yanqi Ge$^{\triangle}$\quad\quad~Ye Huang$^{\triangle}$\thanks{$\triangle$~:~Equal contributions}~\quad\quad~Wen Li\quad\quad~Lixin Duan$^{\Box}$\thanks{$\Box$~: Corresponding author}}
\IEEEauthorblockA{\textit{Shenzhen Institute of Advanced Study, } \\
\textit{University of Electronic Science and Technology of China}}
}
\fi

\maketitle

\begin{abstract}
We introduced SSR, which utilizes SAM (segment-anything) as a strong regularizer during training, to greatly enhance the robustness of the image encoder for handling various domains.
Specifically, given the fact that SAM is pre-trained with a large number of images over the internet, which cover a diverse variety of domains, the feature encoding extracted by the SAM is obviously less dependent on specific domains when compared to the traditional ImageNet pre-trained image encoder.
Meanwhile, the ImageNet pre-trained image encoder is still a mature choice of backbone for the semantic segmentation task, especially when the SAM is category-irrelevant.
As a result, our SSR provides a simple yet highly effective design. 
It uses the ImageNet pre-trained image encoder as the backbone, and the intermediate feature of each stage (\ie there are 4 stages in MiT-B5) is regularized by SAM during training.
After extensive experimentation on GTA5$\rightarrow$Cityscapes, our SSR significantly improved performance over the baseline without introducing any extra inference overhead.

\end{abstract}


\section{Introduction}
Image semantic segmentation is a challenging task in computer vision. 
This technique classifies each pixel of an image and serves as a foundation for visual representation learning.
Since the introduction of Fully Convolutional Networks (FCNs~\cite{cFCN}) and Residual Networks (ResNets~\cite{cResnet}), supervised deep learning models~\cite{cUNet,cDeepLabv1,cDeepLab,cDeepLabV3,cPSPNet,cDeepLabV3Plus,cNonLocal,cDualAttention,cDenseASPP,cExFuse,cCCNet,cANNN,cOCNet,cCFNet,cCPN,cKSAC,cA2Net,cACFNet,cEMANet,cOCR,cDPT,cMaskFormer,cDecoupleSegNet,cGSCNN,cCAA,cCAR,cMask2Former,cDDP,cSegDeformer,cSAR} have achieved remarkable success in the field of semantic segmentation on traditional benchmark datasets (\eg VOC2012, Cityscapes and COCOStuff).

\begin{figure}[t]
\centering
\includegraphics[width=1.0\linewidth]{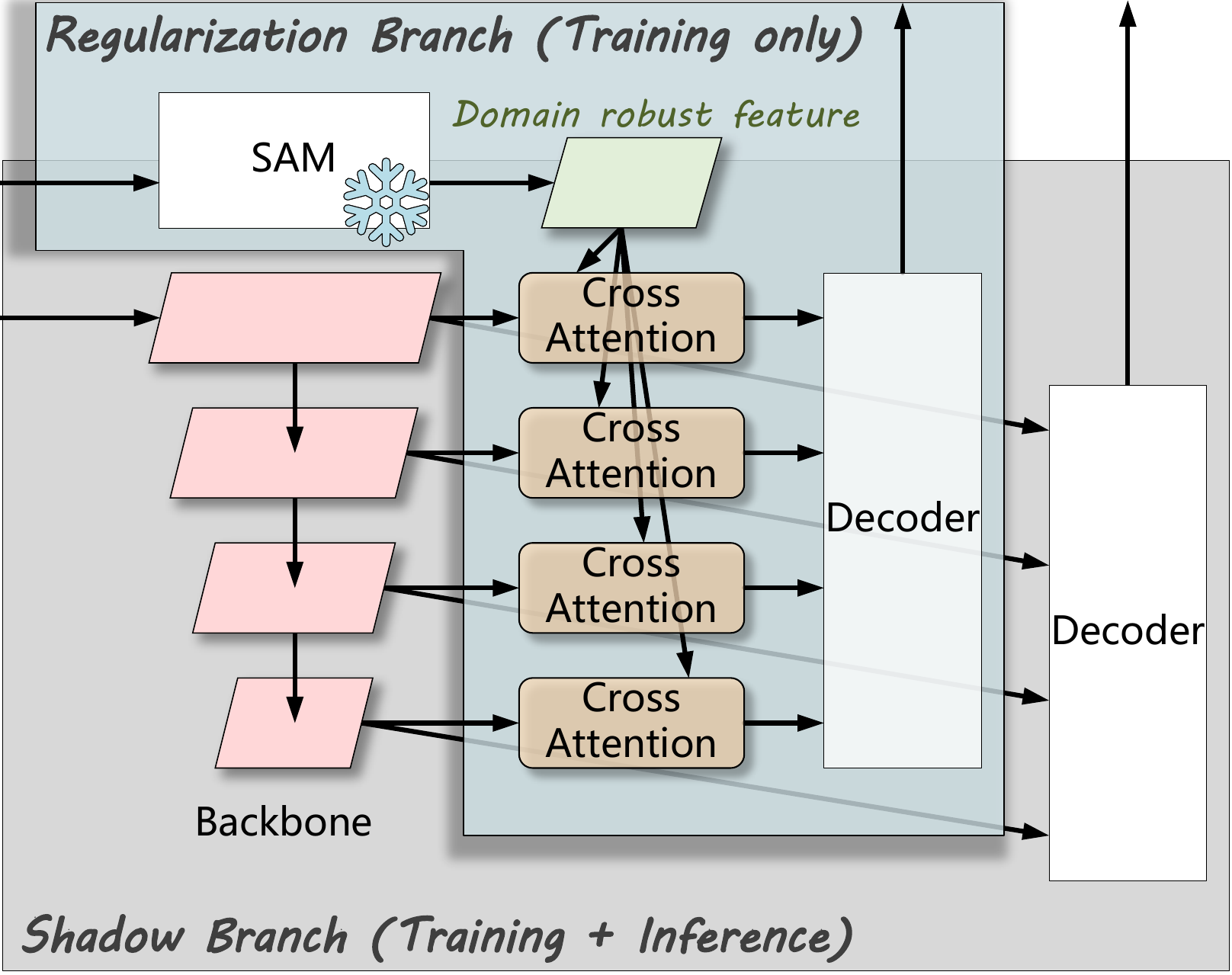}
\caption{
The architecture we propose for our SSR (SAM is a Strong Regularizer) consists of two branches. The Regularization Branch, which is used only during training, includes the frozen SAM. Additionally, we have a simple encoder-decoder shadow branch that is utilized in both the training and inference branches. Zoom in to see better.
}
\label{fig:SSR-SSR}
\end{figure}

In supervised learning, each pixel in an image needs to be labeled, which is also known as dense labeling. This makes semantic segmentation a costly process. For instance, labeling a Cityscapes image can take up to 2 hours for a professional. This high labeling cost leads to limited training data and a lack of diversity, which in turn causes the model to be only generalized on a small domain and has a large domain gap with many others. Ultimately, this affects the robustness of the semantic segmentation model.

Unsupervised Domain Adaptation (UDA) is a research field that aims to tackle the problem of domain gap. Current UDA methods typically focus on improving the model or using more advanced data augmentation strategies. Despite significant progress in recent years, their performance is still noticeably inferior to that of purely supervised models.

In this work, we were inspired by some of the latest ideas in the field, such as the foundation model and big data. We didn't want to limit ourselves to existing UDA patterns, so we decided to use big data to explore new ways of improving UDA performance.
Due to budget constraints, we utilize the foundation model to benefit from internet data indirectly.

SAM (Segment-anything~\cite{cSAM}) is a vision foundation model trained on a massive-scale dataset (SA-1B). The feature map that SAM produced, though it cannot directly make pixel classification, is obviously much more domain-independent than a traditional pixel-segmentation model trained on a small dataset.
Thus, We proposed SSR (SAM is a Strong regularize), using the relatively robust SAM feature map to regularize our model during training to learn much more domain-independent weights for UDA, and during inference, SAM is no longer required for the efficient purpose. Thus, the proposed SSR does not cause extra inference overhead. Experimentally, after the SAM intervention during training, our model's performance significantly improved compared to the baseline.

To summarize, our contributions are listed below:
\begin{itemize}
    \item We use SAM to indirectly take advantage of big data during training and significantly improve the model performance on UDA.
    \item We have designed a `shadow branch` parallel to the SAM regularization so that SAM is not required for inference and avoids extra inference overhead. 
\end{itemize}

\section{Proposed methods}

Our proposed SSR is based on DAFormer, which utilizes MiT-B5 as its backbone and consists of two branches: a regularization branch for training only and a shadow branch for both training and inference.

\subsection{Regularization branch}
SSR froze the weights of SAM to make it under inference mode.
When we input an image, we pass it through both SAM (Segment-Anything) and MiT-B5~\cite{cSegFormer}. 
SAM produces one feature map (see \textcolor{ForestGreen}{green} feature map in Fig.~\ref{fig:SSR-SSR}), while MiT-B5 produces four feature maps (see \textcolor{pink}{pink} feature maps in Fig.~\ref{fig:SSR-SSR}) from its pyramid structure, which consists of four stages.

As SAM is trained on a massive-scale dataset (SA-1B~\cite{cSAM}) that covers diverse domains for each category, the feature map it generates is naturally less dependent on specific domains. Hence, we directly use the SAM feature map to perform the cross-attention with 4 stages outputs of MiT-B5.

In each stage's cross-attention process, the proposed SSR utilizes the MiT-B5 backbone feature map as a query. To align the number of channels with the backbone feature, SSR applies a single linear layer to project the SAM feature map. Since there are four stages of outputs from the MiT-B5 backbone, SSR has four linear layers for channel projection. 

After conducting the cross-attention computation, including the residual add, all of the outputs are passed to the decoder for the rest of the process.

During training, in order to minimize the loss, the cross-attention process ensures the backbone feature map has a similar distribution as the SAM feature map (i.e., the same object has a similar representation). 
In simpler terms, it helps the model learn strong, robust, and adaptable weights that are not limited to a specific domain. This allows a more robust representation of objects, which is crucial for accurate and effective domain adaptive semantic segmentation.

\subsection{Shadow branch}
In addition to the regularization branch, SSR also carries out a shadow branch that operates under the premise that the SAM only serves as a feature regularizer. The shadow branch is solely based on the segmentation model, which shares the backbone and decoder with the regularization branch and does not incorporate the SAM or cross-attention mechanism.

During inference, the shadow branch is the only branch that involves inference, resulting in zero extra inference overhead compared to the DAFormer baseline.

\section{Training details}

We evaluate SSR based on MIC~\cite{hoyer2023mic} and DAFormer~\cite{hoyer2022daformer} with a MiT-B5 encoder~\cite{cSegFormer}. We follow the same network architecture, data augmentation, and optimization technology with DAFormer and MIC.
We use the pre-trained SAM (ViT-b) encoder to regularize the segmentation model.
The experiment is conducted on a common synthetic-to-real benchmark, GTA5$\rightarrow$Cityscapes. The GTA5 dataset~\cite{richter2016playing} is collected from an open-world action game. The Cityscapes~\cite{cCityScapes} is a real-world urban dataset from European cities. 
\section{Experiments}

\subsection{Ablation studies}
We conducted ablation studies in Tab.~\ref{tab:abl1} based on DAFormer, gradually adding SAM-regularized cross-attention to each backbone output stage. As shown in Tab.~\ref{tab:abl1}, we found that regularizing all stages resulted in the highest improvement.

\subsection{Compare with baselines}

We then compare our proposed SSR with DAFormer and MIC on GTA5$\rightarrow$Cityscapes, respectively.
As presented in Tab.~\ref{tab: exp1}, both DAFormer and MIC exhibit improved performance (68.3\% vs. 69.3\% and 70.1\% vs. 70.6\%, respectively) after adding SSR, indicating that our proposed SSR can be generalized across multiple existing approaches.

\begin{table}[t]
\centering
\caption{Ablation studies of applying SAM regularization to different stages of the backbone feature map.}
\resizebox{0.6\linewidth}{!}{
\begin{tabular}{cccc|cc}
\toprule
 S0 & S1  & S2 & S3 & mIoU(\%) & $\Delta$ \\ 
\midrule
  & & &  &  68.3 & - \\
  
  & & & \checkmark &  68.8 & 0.5 $\uparrow$\\
  & &\checkmark  & \checkmark & 69.0 & 0.7 $\uparrow$\\
  &\checkmark &\checkmark  & \checkmark & 68.7 & 0.4 $\uparrow$\\
 % \checkmark  &  &\checkmark  & \checkmark & 68.8\\
    \checkmark&\checkmark &\checkmark  & \checkmark & \textbf{69.3} & \textbf{1.0} $\uparrow$\\
\bottomrule

\end{tabular}
}
% model:DAFormer, SAM: vit-b.
\label{tab:abl1} 
\end{table}

\begin{table*}[h]
\centering
\caption{Comparison results of GTA5$\rightarrow$Cityscapes adaptation task. }
\label{tab: exp1}
\setlength{\tabcolsep}{3pt}
\resizebox{\textwidth}{!}{%
\begin{tabular}{l|ccccccccccccccccccc|c}
\toprule
 & Road & S.walk & Build. & Wall & Fence & Pole & Tr.Light & Sign & Veget. & Terrain & Sky & Person & Rider & Car & Truck & Bus & Train & M.bike & Bike & mIoU(\%)\\
\midrule

DACS~\cite{tranheden2021dacs} & 95.0 & 68.3 & 87.6 & 36.8 & 35.0 & 37.4 & 50.0 & 53.2 & 88.3 & 46.0 & 88.9 & 68.7 & 43.1 & 88.1 & 50.7 & 54.2 & 1.0 & 47.8 & 57.2 & 57.8\\

Undoing~\cite{liu2022undoing} &92.9 &52.7& 87.2 &39.4&41.3& 43.9& 55.0 &52.9 & 89.3 &48.2 & 91.2 &71.4 & 36.0 &90.2 & 67.9 &59.8 & 0.0 &48.5 & 59.3 & 59.3\\ 

DAFormer~\cite{hoyer2022daformer} &95.7 & 70.2 & 89.4 & 53.5 & 48.1 & 49.6 & 55.8 & 59.4 & 89.9 & 47.9 & 92.5 & 72.2 & 44.7 & 92.3 & 74.5 & 78.2 & 65.1 & 55.9 & 61.8 & 68.3\\

MIC~\cite{hoyer2023mic} &  96.9& 76.2& \textbf{90.2} & \textbf{59.1}& 50.7& \textbf{52.8}& 55.6& 60.5& \textbf{90.0}& \textbf{49.0} & \textbf{92.8}&\textbf{73.2}& \textbf{48.0} & 92.6 & 75.8& \textbf{83.7}& 69.4& 54.1& 62.7 & 70.1\\

\midrule
DAFormer + SSR & 96.5 & 73.2& 89.4& 54.4 & 43.3 & 50.4 & \textbf{56.5} & 61.2 & \textbf{90.0} & 45.4 & \textbf{92.8}& 72.4& 45.8& 93.0& 80.4& 79.3 & 70.5 & \textbf{58.4} & \textbf{64.7} & 69.3 \\

MIC + SSR & \textbf{97.3} & \textbf{78.2} & \textbf{90.2} & 58.1& \textbf{51.9} & \textbf{52.8}& 56.2& \textbf{64.6} & 89.8& 46.7& 91.9& 72.7 & 45.6& \textbf{92.9}& \textbf{82.5}& 78.8& \textbf{72.0}& 56.1& 63.8& \textbf{70.6} \\
\bottomrule
\end{tabular}
}
\end{table*}

\section{Visualization}

\begin{figure}[t]
\centering
\includegraphics[width=1.0\linewidth]{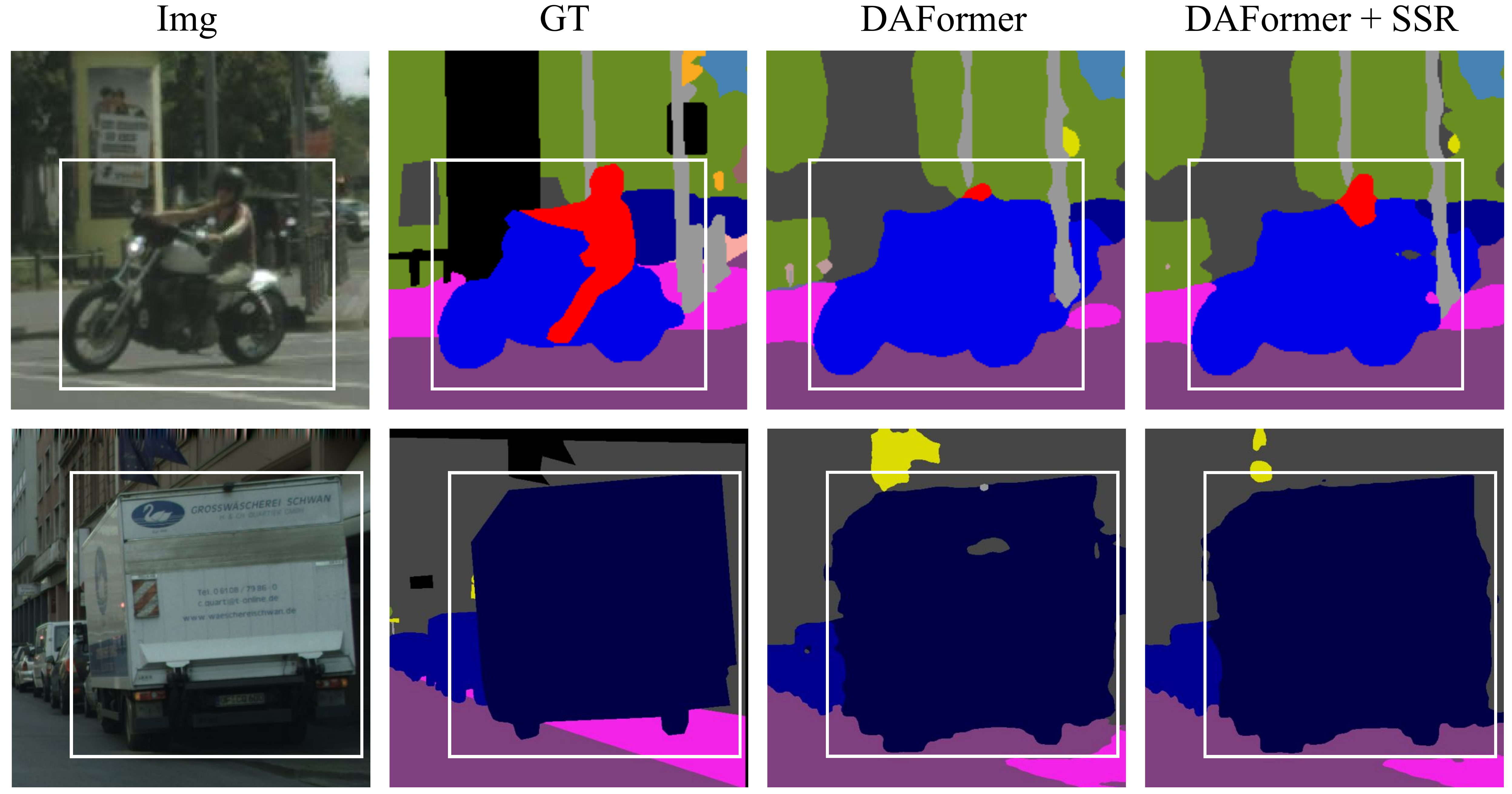}
\caption{
Comparison of DAFormer vs DAFormer + SSR on Cityscapes dataset.
}
\label{fig:SSR-vis}
\end{figure}

Fig.~\ref{fig:SSR-vis} provides visualizations for DAFormer and DAFormer + SSR. The visualization shows that using SAM as a regularizer during training reduces misclassification.

\section{Conclusion}

In this work, we introduce SSR (SAM is a Strong Regularizer), a new strategy to enhance the performance of domain adaptive semantic segmentation. SAM, which was trained on a massive-scale dataset, effectively regularizes the learning process of the model and enables it to produce more domain-independent representations. The shadow branch, which shares the same model, ensures that SSR has zero extra inference overhead. An ablation study and experiment are carried out to validate the effectiveness of the proposed SSR.

\bibliographystyle{IEEEtran}

\begin{thebibliography}{10}
\providecommand{\url}[1]{#1}
\csname url@samestyle\endcsname
\providecommand{\newblock}{\relax}
\providecommand{\bibinfo}[2]{#2}
\providecommand{\BIBentrySTDinterwordspacing}{\spaceskip=0pt\relax}
\providecommand{\BIBentryALTinterwordstretchfactor}{4}
\providecommand{\BIBentryALTinterwordspacing}{\spaceskip=\fontdimen2\font plus
\BIBentryALTinterwordstretchfactor\fontdimen3\font minus \fontdimen4\font\relax}
\providecommand{\BIBforeignlanguage}[2]{{%
\expandafter\ifx\csname l@#1\endcsname\relax
\typeout{** WARNING: IEEEtran.bst: No hyphenation pattern has been}%
\typeout{** loaded for the language `#1'. Using the pattern for}%
\typeout{** the default language instead.}%
\else
\language=\csname l@#1\endcsname
\fi
#2}}
\providecommand{\BIBdecl}{\relax}
\BIBdecl

\bibitem{cFCN}
J.~Long, E.~Shelhamer, and T.~Darrell, ``Fully convolutional networks for semantic segmentation,'' in \emph{IEEE Conference on Computer Vision and Pattern Recognition}, 2015.

\bibitem{cResnet}
K.~He, X.~Zhang, S.~Ren, and J.~Sun, ``Deep residual learning for image recognition,'' in \emph{IEEE Conference on Computer Vision and Pattern Recognition}, 2016.

\bibitem{cUNet}
O.~Ronneberger, P.~Fischer, and T.~Brox, ``U-net: Convolutional networks for biomedical image segmentation,'' in \emph{MICCAI}, 2015.

\bibitem{cDeepLabv1}
L.-C. Chen, G.~Papandreou, I.~Kokkinos, K.~Murphy, and A.~L. Yuille, ``Semantic image segmentation with deep convolutional nets and fully connected crfs,'' in \emph{International Conference on Learning Representations}, 2014.

\bibitem{cDeepLab}
------, ``Deeplab: Semantic image segmentation with deep convolutional nets, atrous convolution, and fully connected crfs,'' \emph{IEEE Transactions on Pattern Analysis and Machine Intelligence}, 2017.

\bibitem{cDeepLabV3}
L.-C. Chen, G.~Papandreou, F.~Schroff, and H.~Adam, ``Rethinking atrous convolution for semantic image segmentation,'' 2017.

\bibitem{cPSPNet}
H.~Zhao, J.~Shi, X.~Qi, X.~Wang, and J.~Jia, ``Pyramid scene parsing network,'' in \emph{IEEE Conference on Computer Vision and Pattern Recognition}, 2017.

\bibitem{cDeepLabV3Plus}
L.-C. Chen, Y.~Zhu, G.~Papandreou, F.~Schroff, and H.~Adam, ``Encoder-decoder with atrous separable convolution for semantic image segmentation,'' in \emph{European Conference on Computer Vision}, 2018.

\bibitem{cNonLocal}
X.~Wang, R.~Girshick, A.~Gupta, and K.~He, ``Non-local neural networks,'' in \emph{IEEE Conference on Computer Vision and Pattern Recognition}, 2018.

\bibitem{cDualAttention}
J.~Fu, J.~Liu, H.~Tian, Y.~Li, Y.~Bao, Z.~Fang, and H.~Lu, ``Dual attention network for scene segmentation,'' in \emph{IEEE Conference on Computer Vision and Pattern Recognition}, 2019.

\bibitem{cDenseASPP}
M.~Yang, K.~Yu, C.~Zhang, Z.~Li, and K.~Yang, ``Denseaspp for semantic segmentation in street scenes,'' in \emph{IEEE Conference on Computer Vision and Pattern Recognition}, 2018.

\bibitem{cExFuse}
Z.~Zhang, X.~Zhang, C.~Peng, X.~Xue, and J.~Sun, ``Exfuse: enhancing feature fusion for semantic segmentation,'' in \emph{European Conference on Computer Vision}, 2018.

\bibitem{cCCNet}
Z.~Huang, X.~Wang, Y.~Wei, L.~Huang, H.~Shi, W.~Liu, and T.~S. Huang, ``Ccnet: Criss-cross attention for semantic segmentation,'' \emph{IEEE Transactions on Pattern Analysis and Machine Intelligence}, 2020.

\bibitem{cANNN}
Z.~Zhu, M.~Xu, S.~Bai, T.~Huang, and X.~Bai, ``Asymmetric non-local neural networks for semantic segmentation,'' in \emph{International Conference on Computer Vision}, 2019.

\bibitem{cOCNet}
Y.~Yuan, L.~Huang, J.~Guo, C.~Zhang, X.~Chen, and J.~Wang, ``Ocnet: Object context network for scene parsing,'' \emph{International Journal of Computer Vision}, 2021.

\bibitem{cCFNet}
H.~Zhang, H.~Zhan, C.~Wang, and J.~Xie, ``Semantic correlation promoted shape-variant context for segmentation,'' in \emph{IEEE Conference on Computer Vision and Pattern Recognition}, 2019.

\bibitem{cCPN}
C.~Yu, J.~Wang, C.~Gao, G.~Yu, C.~Shen, and N.~Sang, ``Context prior for scene segmentation,'' in \emph{IEEE Conference on Computer Vision and Pattern Recognition}, 2020.

\bibitem{cKSAC}
Y.~Huang, Q.~Wang, W.~Jia, and X.~He, ``See more than once--kernel-sharing atrous convolution for semantic segmentation,'' \emph{Neuro Computing}, 2021.

\bibitem{cA2Net}
Y.~Chen, Y.~Kalantidis, J.~Li, S.~Yan, and J.~Feng, ``A2-nets: Double attention networks,'' in \emph{Conference on Neural Information Processing Systems}, 2018.

\bibitem{cACFNet}
F.~Zhang, Y.~Chen, Z.~Li, Z.~Hong, J.~Liu, F.~Ma, J.~Han, and E.~Ding, ``Acfnet: Attentional class feature network for semantic segmentation,'' in \emph{International Conference on Computer Vision}, 2019.

\bibitem{cEMANet}
X.~Li, Z.~Zhong, J.~Wu, Y.~Yang, Z.~Lin, and H.~Liu, ``Expectation-maximization attention networks for semantic segmentation,'' in \emph{International Conference on Computer Vision}, 2019.

\bibitem{cOCR}
Y.~Yuan, X.~Chen, and J.~Wang, ``Object-contextual representations for semantic segmentation,'' in \emph{European Conference on Computer Vision}, 2020.

\bibitem{cDPT}
R.~Ranftl, A.~Bochkovskiy, and V.~Koltun, ``Vision transformers for dense prediction,'' in \emph{ICCV}, 2021.

\bibitem{cMaskFormer}
B.~Cheng, A.~G. Schwing, and A.~Kirillov, ``Per-pixel classification is not all you need for semantic segmentation,'' in \emph{Conference on Neural Information Processing Systems}, 2021.

\bibitem{cDecoupleSegNet}
X.~Li, X.~Li, L.~Zhang, C.~Guangliang, J.~Shi, Z.~Lin, Y.~Tong, and S.~Tan, ``Improving semantic segmentation via decoupled body and edge supervision,'' in \emph{European Conference on Computer Vision}, 2020.

\bibitem{cGSCNN}
T.~Takikawa, D.~Acuna, V.~Jampani, and S.~Fidler, ``Gated-scnn: Gated shape cnns for semantic segmentation,'' in \emph{International Conference on Computer Vision}, 2019.

\bibitem{cCAA}
Y.~Huang, D.~Kang, W.~Jia, X.~He, and L.~liu, ``Channelized axial attention - considering channel relation within spatial attention for semantic segmentation,'' in \emph{AAAI}, 2022.

\bibitem{cCAR}
Y.~Huang, D.~Kang, L.~Chen, X.~Zhe, W.~Jia, L.~Bao, and X.~He, ``Car: Class-aware regularizations for semantic segmentation,'' in \emph{European Conference on Computer Vision}, 2022.

\bibitem{cMask2Former}
B.~Cheng, I.~Misra, A.~G. Schwing, A.~Kirillov, and R.~Girdhar, ``Masked-attention mask transformer for universal image segmentation,'' in \emph{IEEE Conference on Computer Vision and Pattern Recognition}, 2022.

\bibitem{cDDP}
Y.~Ji, Z.~Chen, E.~Xie, L.~Hong, X.~Liu, Z.~Liu, T.~Lu, Z.~Li, and P.~Luo, ``Ddp: Diffusion model for dense visual predition,'' in \emph{International Conference on Computer Vision}, 2023.

\bibitem{cSegDeformer}
B.~Shi, D.~Jiang, X.~Zhang, H.~Li, W.~Dai, J.~Zou, H.~Xiong, and Q.~Tian, ``A transformer-based decoder for semantic segmentation with multi-level context mining,'' in \emph{European Conference on Computer Vision}, 2022.

\bibitem{cSAR}
Y.~Ge, Q.~Nie, Y.~Huang, Y.~Liu, C.~Wang, F.~Zheng, W.~Li, and L.~Duan, ``Beyond prototypes: Semantic anchor regularization for better representation learning,'' in \emph{AAAI}, 2024.

\bibitem{cSAM}
A.~Kirillov, E.~Mintun, N.~Ravi, H.~Mao, C.~Rolland, L.~Gustafson, T.~Xiao, S.~Whitehead, A.~C. Berg, W.-Y. Lo, P.~Dollár, and R.~Girshick, ``Segment anything,'' in \emph{International Conference on Computer Vision}, 2023.

\bibitem{cSegFormer}
E.~Xie, W.~Wang, Z.~Yu, A.~Anandkumar, J.~M. Alvarez, and P.~Luo, ``Segformer: Simple and efficient design for semantic segmentation with transformers,'' in \emph{Conference on Neural Information Processing Systems}, 2021.

\bibitem{hoyer2023mic}
L.~Hoyer, D.~Dai, H.~Wang, and L.~Van~Gool, ``{MIC}: Masked image consistency for context-enhanced domain adaptation,'' in \emph{IEEE Conference on Computer Vision and Pattern Recognition}, 2023.

\bibitem{hoyer2022daformer}
L.~Hoyer, D.~Dai, and L.~Van~Gool, ``Daformer: Improving network architectures and training strategies for domain-adaptive semantic segmentation,'' in \emph{IEEE Conference on Computer Vision and Pattern Recognition}, 2022, pp. 9924--9935.

\bibitem{richter2016playing}
S.~R. Richter, V.~Vineet, S.~Roth, and V.~Koltun, ``Playing for data: Ground truth from computer games,'' in \emph{Computer Vision--ECCV 2016: 14th European Conference, Amsterdam, The Netherlands, October 11-14, 2016, Proceedings, Part II 14}.\hskip 1em plus 0.5em minus 0.4em\relax Springer, 2016, pp. 102--118.

\bibitem{cCityScapes}
C.~Marius, O.~Mohamed, R.~Sebastian, R.~Timo, E.~Markus, B.~Rodrigo, F.~Uwe, S.~Roth, and S.~Bernt, ``The cityscapes dataset for semantic urban scene understanding,'' in \emph{IEEE Conference on Computer Vision and Pattern Recognition}, 2016.

\bibitem{tranheden2021dacs}
W.~Tranheden, V.~Olsson, J.~Pinto, and L.~Svensson, ``Dacs: Domain adaptation via cross-domain mixed sampling,'' in \emph{IEEE Conference on Computer Vision and Pattern Recognition}, 2021, pp. 1379--1389.

\bibitem{liu2022undoing}
Y.~Liu, J.~Deng, J.~Tao, T.~Chu, L.~Duan, and W.~Li, ``Undoing the damage of label shift for cross-domain semantic segmentation,'' in \emph{IEEE Conference on Computer Vision and Pattern Recognition}, 2022, pp. 7042--7052.

\end{thebibliography}

\clearpage

\end{document}